\documentclass[10pt,twocolumn,letterpaper]{article}

\usepackage{cvpr}              

\usepackage{graphicx}
\usepackage{amsmath}
\usepackage{amssymb}
\usepackage{booktabs}
\usepackage{multicol,multirow}
\usepackage{setspace}
\usepackage{makecell}
\usepackage[dvipsnames]{xcolor}

\usepackage[pagebackref,breaklinks,colorlinks]{hyperref}

\usepackage[capitalize]{cleveref}
\crefname{section}{Sec.}{Secs.}
\Crefname{section}{Section}{Sections}
\Crefname{table}{Table}{Tables}
\crefname{table}{Tab.}{Tabs.}

\def\cvprPaperID{5505} 
\def\confName{CVPR}
\def\confYear{2023}

\begin{document}

\title{MonoATT: Online Monocular 3D Object Detection \\with Adaptive Token Transformer}

\author{
Yunsong Zhou \textsuperscript{\rm 1}~~ 
Hongzi Zhu \textsuperscript{\rm 1}
\thanks{Corresponding authors}
~~ 
Quan Liu \textsuperscript{\rm 1}~~
Shan Chang \textsuperscript{\rm 2}~~ 
Minyi Guo \textsuperscript{\rm 1}~~  \\
\textsuperscript{\rm 1}Shanghai Jiao Tong University~~
\textsuperscript{\rm 2}Donghua University
\\
\text{\{zhouyunsong,hongzi,liuquan2017,guo-my\}@sjtu.edu.cn}~~
\text{changshan@dhu.edu.cn}
}
\maketitle

\begin{abstract}
Mobile monocular 3D object detection (Mono3D) (e.g., on a vehicle, a drone, or a robot) is an important yet challenging task. 
Existing transformer-based offline Mono3D models adopt grid-based vision tokens, which is suboptimal when using coarse tokens due to the limited available computational power.
In this paper, we propose an online Mono3D framework, called \emph{MonoATT}, which leverages a novel vision transformer with heterogeneous tokens of varying shapes and sizes to facilitate mobile Mono3D. The core idea of MonoATT is to adaptively assign finer tokens to areas of more significance before utilizing a transformer to enhance Mono3D. To this end, we first use prior knowledge to design a scoring network for selecting the most important areas of the image, and then propose a token clustering and merging network with an attention mechanism to gradually merge tokens around the selected areas in multiple stages. Finally, a pixel-level feature map is reconstructed from heterogeneous tokens before employing a SOTA Mono3D detector as the underlying detection core. 
Experiment results on the real-world KITTI dataset demonstrate that MonoATT can effectively improve the Mono3D accuracy for both near and far objects and guarantee low latency. 
MonoATT yields the best performance compared with the state-of-the-art methods by a large margin and is ranked number one on the KITTI 3D benchmark.
\end{abstract}
\vspace{-0.7cm}

\section{Introduction}
\label{sec:intro}
\vspace{-0.25cm}

\begin{figure}
    \centering
    \includegraphics[width=0.95\linewidth]{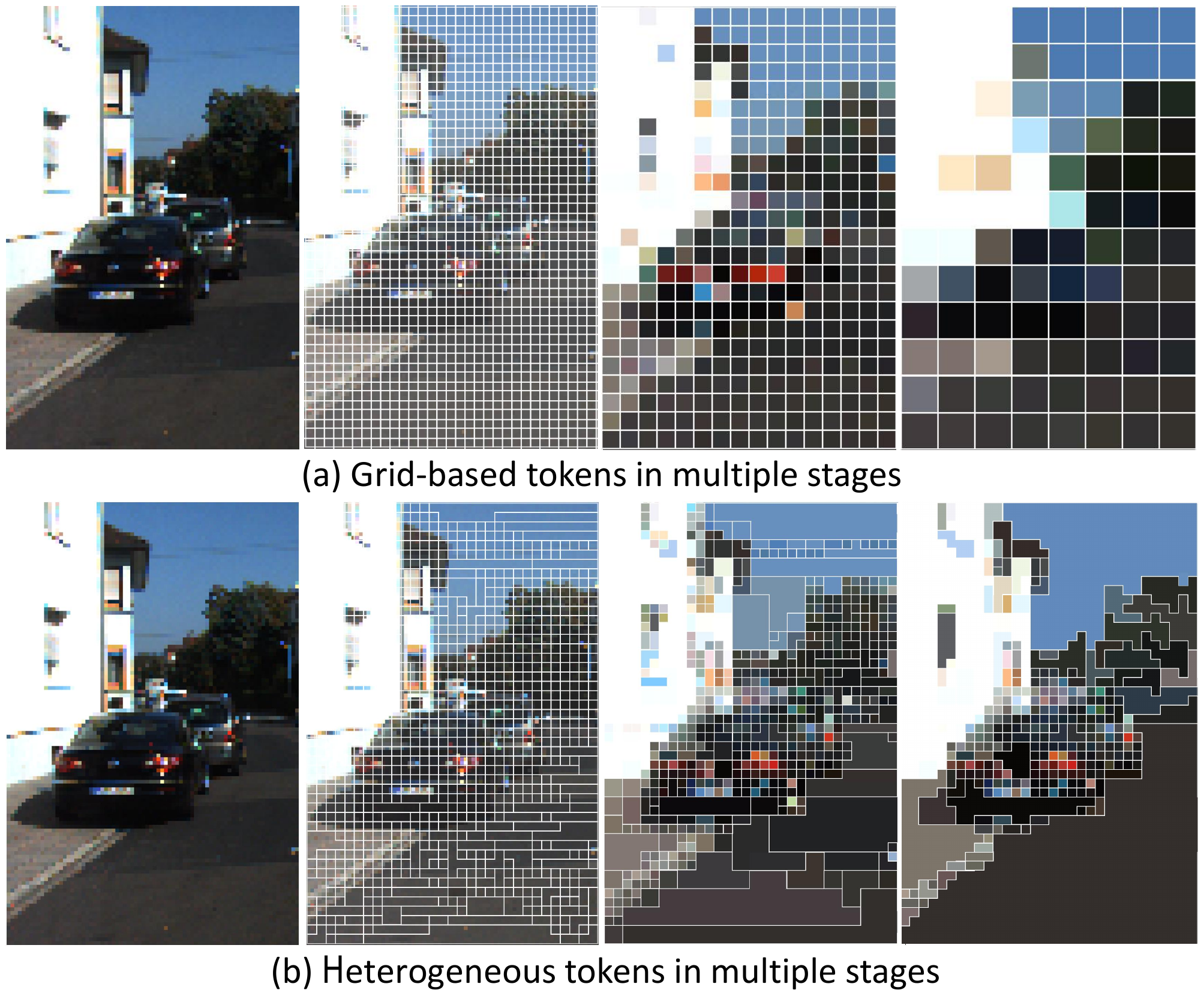}
    \vspace{-0.3cm}
    \caption{Illustration of (a) grid-based tokens used in traditional vision transformers and (b) heterogeneous tokens used in our adaptive token transformer (ATT). Instead of equally treating all image regions, our ATT distributes dense and fine tokens to meaningful image regions (\textit{i.e.}, distant cars and lane lines) yet coarse tokens to regions with less information such as the background.
    }
    \label{fig:intro}
    \vspace{-0.8cm}
\end{figure}

Three-dimensional (3D) object detection has long been a fundamental problem in both industry and academia and enables various applications, ranging from autonomous vehicles \cite{Geiger2012CVPR} and drones, to robotic manipulation and augmented reality applications.
Previous methods have achieved superior performance based on the accurate depth information from multiple sensors, such as LiDAR signal \cite{chen_multiview_2016, qi_frustum_2017, shi_pointrcnn_2018, shin_roarnet_2018, liang_deep_2018, zhou_voxelnet_2017} or stereo matching \cite{chen_3d_2015, chen_3d_2018, li_stereo_2019, pham_robust_2017, qin_triangulation_2019, xu_multilevel_2018}.
In order to lower the sensor requirements, a much cheaper, more energy-efficient, and easier-to-deploy alternative, \textit{i.e.}, monocular 3D object detection (Mono3D) has been proposed and made impressive progress.
A practical online Mono3D detector for autonomous driving should meet the following two requirements: 1) given the constrained computational resource on a mobile platform, the 3D bounding boxes produced by the Mono3D detector should be accurate enough, not only for near objects but also for far ones, to ensure, \textit{e.g.}, high-priority driving safety applications; 2) the response time of the Mono3D detector should be as low as possible to ensure that objects of interest can be instantly detected in mobile settings.

Current Mono3D methods, such as depth map based \cite{qin_monogrnet_2018, ma2020rethinking, ding2020learning}, pseudo-LiDAR based \cite{ma_accurate_2019, manhardt_roi_10d_2018, wang_pseudo-lidar_2018, xu_multilevel_2018,qin_monogrnet_2018, ding2020learning, ma2020rethinking}, and image-only based \cite{ren2015faster,tian2019fcos,zhou_objects_2019,brazil_m3d_rpn_2019, liu2020smoke, wang2021fcos3d,chen2020monopair,li2020rtm3d,brazil2020kinematic,zhou2021monocular,zhang2021objects,lu2021geometry,zhang2022monodetr}, mostly follow the pipelines of traditional 2D object detectors \cite{ren2015faster, redmon2018yolov3, zhou_objects_2019, tian2019fcos} to first localize object centers from heatmaps and then aggregate visual features around each object center to predict the object's 3D properties, \textit{e.g.}, location, depth, 3D sizes, and orientation.
Although it is conceptually straightforward and has low computational overhead, merely using local features around the predicted object centers is insufficient to understand the scene-level geometric cues for accurately estimating the depth of objects, making existing Mono3D methods far from satisfactory.
Recently, inspired by the success of transformers in natural language processing, visual transformers with long-range attention between image patches have recently been developed to solve Mono3D tasks and achieve state-of-the-art (SOTA) performance \cite{huang2022monodtr,zhang2022monodetr}. As illustrated in Figure \ref{fig:intro} (a), most existing vision transformers follow the \textit{grid-based} token generation method, where an input image is divided into a grid of equal image patches, known as tokens.
However, using grid-based tokens is sub-optimal for Mono3D applications such as autonomous driving because of the following two reasons: 1) far objects have smaller size and less image information, which makes them hard to detect with coarse grid-based tokens; 2) using fine grid-based tokens is prohibitive due to the limited computational power and the stringent latency requirement.

In this paper, we propose an online Mono3D framework, called \emph{MonoATT}, which leverages a novel vision transformer with \emph{heterogeneous} tokens of varying sizes and shapes to boost mobile Mono3D. We have one key observation that not all image pixels of an object have equivalent significance with respect to Mono3D. For instance, pixels on the outline of a vehicle are more important than those on the body; pixels on far objects are more sensitive than those on near objects. The core idea of MonoATT is to automatically assign fine tokens to pixels of more significance and coarse tokens to pixels of less significance before utilizing a transformer to enhance Mono3D detection. To this end, as illustrated in Figure \ref{fig:intro} (b), we apply a \textit{similarity compatibility principle} to dynamically cluster and aggregate image patches with similar features into heterogeneous tokens in multiple stages. In this way, MonoATT neatly distributes computational power among image parts of different importance, satisfying both the high accuracy and low response time requirements posed by mobile Mono3D applications.

There are three main challenges in designing MonoATT.
First, it is essential yet non-trivial to determine keypoints on the feature map which can represent the most relevant information for Mono3D detection. Such keypoints also serve as cluster centers to group tokens with similar features.
To tackle this challenge, we score image features based on prior knowledge in mobile Mono3D scenarios. Specifically, features of targets (\emph{e.g.}, vehicles, cyclists, and pedestrians) are more important than features of the background. Moreover, more attention is paid to features of distant targets and the outline of targets. Then, a predefined number of keypoints with the highest scores are selected as cluster centers to guide the token clustering in each stage. 
As a result, an image region with dense keypoints will eventually be assigned with fine tokens while a region with sparse keypoints will be assigned with coarse tokens. 

Second, given the established cluster centers in each stage, how to group similar tokens into clusters and effectively aggregate token features within a cluster is non-intuitive.
Due to the local correlation of 2D convolution, using naive minimal feature distance for token clustering would make the model insensitive to object outlines. Furthermore, a straightforward feature averaging scheme would be greatly affected by noise introduced by outlier tokens.
To deal with these issues, we devise a token clustering and merging network. It groups tokens into clusters, taking both the feature similarity and image distance between tokens into account, so that far tokens with similar features are more likely to be designated into one cluster. Then, it merges all tokens in a cluster into one combined token and aggregates their features with an attention mechanism.

Third, recovering multi-stage vision tokens to a pixel-level feature map is proved to be beneficial for vision transformers \cite{Sun_2019_CVPR, YuanFHLZCW21}. However, how to restore a regular image feature map from heterogeneous tokens of irregular shapes and various sizes is challenging.
To transform adaptive tokens of each stage into feature maps, we propose an efficient multi-stage feature reconstruction network.
Specifically, the feature reconstruction network starts from the last stage of clustering, gradually upsamples the tokens, and aggregates the token features of the previous stage.
The aggregated tokens correspond to the pixels in the feature map one by one and are reshaped into a feature map.
As a result, accurate 3D detection results can be obtained via a conventional Mono3D detector using the enhanced feature map.

Experiments on KITTI dataset \cite{Geiger2012CVPR} demonstrate that our method outperforms the SOTA methods by a large margin. 
Such a framework can be applied to existing Mono3D detectors and is practical for industrial applications. 
The proposed MonoATT is ranked \textit{number one} on the KITTI 3D benchmark by submission. The whole suite of the code base will be released and the experimental results will be posted to the public leaderboard. We highlight the main contributions made in this paper as follows: 
1) a novel online Mono3D framework is introduced, leveraging an adaptive token transformer to improve the detection accuracy and guarantee a low latency;
2) a scoring network is proposed, which integrates prior knowledge to estimate keypoints for progressive adaptive token generation;
3) a feature reconstruction network is designed to reconstruct a detailed image feature map from adaptive tokens efficiently.

\begin{figure*}
    \centering
    \includegraphics[width=0.74\textwidth]{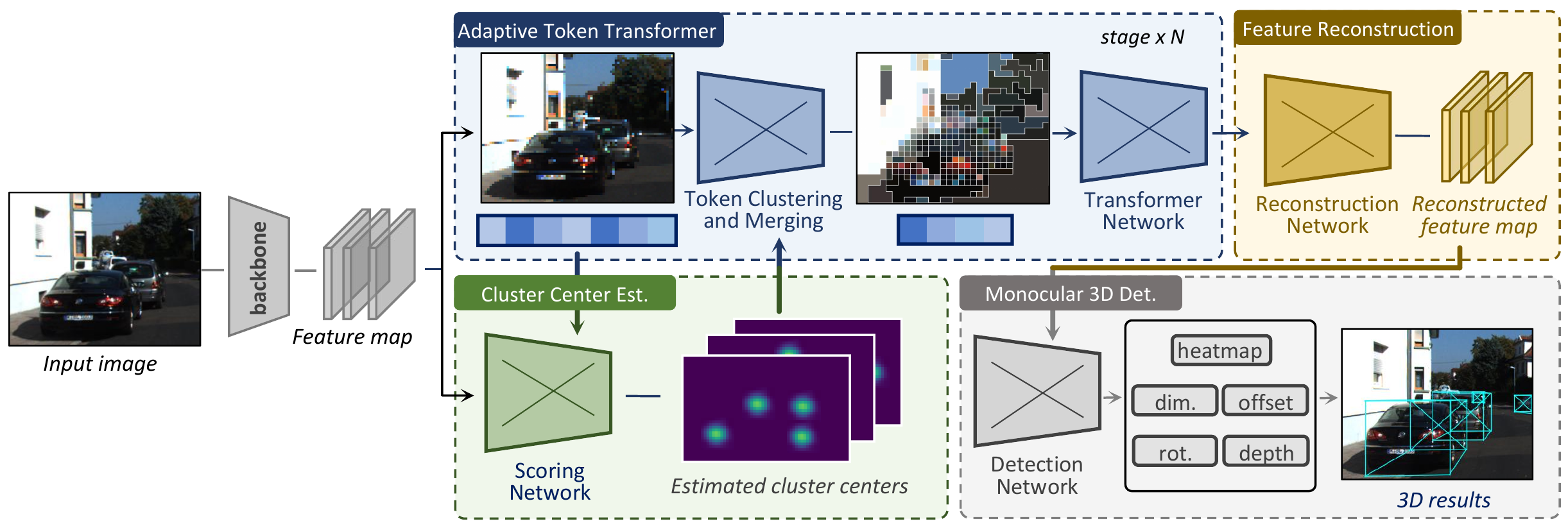}
    \vspace{-0.2cm}
    \caption{MonoATT consists of four main components, \emph{i.e.}, \emph{cluster center estimation} (CCE), \emph{adaptive token transformer} (ATT), \emph{multi-stage feature reconstruction} (MFR), and \emph{monocular 3D detection}. 
CCE involves a scoring network to predict the most essential image areas which serve as cluster centers in each heterogeneous token generation stage. 
Given the initial fine grid-based tokens obtained by slicing the feature map, ATT first generates heterogeneous tokens adaptive to the significance of image areas by grouping and merging tokens in multiple stages; then it leverages the long-range self-attention mechanism provided by a transformer network to associate features on heterogeneous tokens. MFR reconstructs an enhanced pixel-level feature map from all irregular tokens for the ease of Mono3D. Finally, a standard Mono3D detector is employed as the underlying detection core.}
\label{fig:overview}
\vspace{-0.5cm}
\end{figure*}

\vspace{-0.2cm}
\section{Related Work}
\vspace{-0.2cm}

\textbf{Standard Monocular 3D object detection.}
The monocular 3D object detection aims to predict 3D bounding boxes from a single given image.
Except for methods assisted by additional inputs, such as depth maps~\cite{qin_monogrnet_2018, ma2020rethinking, ding2020learning}, CAD models~\cite{chen_monocular_2016,murthy2017reconstructing,chabot2017deep,xiang_subcategory-aware_2017,liu2021autoshape}, and LiDAR~\cite{wang_pseudo-lidar_2018,ma_accurate_2019,chen2021monorun,reading2021categorical}, standard monocular detectors \cite{kundu20183d,he2019mono3d++,liu2019deep,simonelli2019disentangling,zhou2021monoef} take as input only a single image and mostly adopt center-guided pipelines following conventional 2D detectors~\cite{ren2015faster,zhou_objects_2019,tian2019fcos}.
M3D-RPN~\cite{brazil_m3d_rpn_2019} designs a depth-aware convolution along with 3D anchors to generate better 3D region proposals. 
With very few handcrafted modules, SMOKE~\cite{liu2020smoke} and FCOS3D~\cite{wang2021fcos3d} propose concise architectures for one-stage monocular detection built on CenterNet~\cite{zhou_objects_2019} and FCOS~\cite{tian2019fcos}, respectively.
Many methods turn to geometric constraints for improving performance.
MonoPair~\cite{chen2020monopair} considers adjacent object pairs and parses their spatial relations with uncertainty.
MonoEF \cite{zhou2021monocular} first proposes a novel method to capture the camera pose in order to formulate detectors that are not subject to camera extrinsic perturbations.
MonoFlex~\cite{zhang2021objects} conducts an uncertainty-guided depth ensemble and categorizes different objects for distinctive processing. 
GUPNet~\cite{lu2021geometry} solves the error amplification problem by geometry-guided depth uncertainty and collocates a hierarchical learning strategy to reduce the training instability.
To further strengthen the detection accuracy, recent methods have introduced more effective but complicated vision transformers into the networks. 
MonoDTR~\cite{huang2022monodtr} proposes to globally integrate context- and depth-aware features with transformers and inject depth hints into the transformer for better 3D reasoning.
MonoDETR~\cite{zhang2022monodetr} adopts a depth-guided feature aggregation scheme via a depth-guided transformer and discards the dependency for center detection.
The above geometrically dependent designs largely promote the overall performance of image-only methods, but the underlying problem still exists, namely, the detection accuracy for distant objects is still not satisfactory.

\textbf{Object detection via the transformer.}
Transformer \cite{vaswani2017attention} is first introduced in sequential modeling in natural language processing tasks, and it has been successfully leveraged in DETR \cite{carion2020end} which improves the detection performance in the computer vision field by using the long-range attention mechanism.
Several methods \cite{cheng2021swin, yang2021transformer,yang2021transformers} have made a demonstration of how to apply a vision transformer to a monocular camera model.
Transformerfusion \cite{bozic2021transformerfusion} leverages the transformer architecture so that the network learns to focus on the most relevant image frames for each 3D location in the scene, supervised only by the scene reconstruction task. 
MT-SfMLearner \cite{varma2022transformers} first demonstrates how to adapt vision transformers for self-supervised monocular depth estimation focusing on improving the robustness of natural corruptions.
Some recent works, MonoDTR \cite{huang2022monodtr} and MonoDETR \cite{zhang2022monodetr}, have tried to apply transformers to monocular 3D detection tasks.
However, the token splitting of these models is still based on a grid of regular shapes and sizes.
None of these methods consider how to merge unnecessary tokens to reduce the computational complexity of the high-resolution feature maps, which will not be available in a typical Mono3D task in autonomous driving scenarios.

There exist some schemes working on improving the efficiency of transformers.
Yu \textit{et al.} \cite{yu2022k} propose to reformulate the cross-attention learning as a clustering process. 
Some approaches \cite{fayyaz2022adaptive,rao2021dynamicvit,tang2022patch} study the efficiency of ViTs and propose a dynamic token sparsification framework to prune redundant tokens progressively.
Wang \textit{et al.} \cite{wang2021not} propose to automatically configure a proper number of tokens for each input image.
The methods mentioned above are all variants of grid-based token generation, which modify the resolution, centers of grids, and number specifically. 
In contrast, the token regions of MonoATT are not restricted by grid structure and are more flexible in three aspects, \textit{i.e.} location, shape, and size.


Our MonoATT inherits DETR’s superiority for non-local encoding and long-range attention. 
Inspired by transformers based on variant-scaled and sparse tokens \cite{rao2021dynamicvit, wang2021not, zeng2022not}, we use dynamically generated adaptive tokens to obtain high accuracy for both near and far targets with low computational overhead.

\section{Design of MonoATT}
\vspace{-0.2cm}

The guiding philosophy of MonoATT is to utilize adaptive tokens with irregular shapes and various sizes to enhance the representation of image features for transformer-based Mono3D so that two goals can be achieved: 
1) superior image features are obtained from coarse to fine to increase Mono3D accuracy for both near and far objects;
2) irrelevant information (\textit{e.g.}, background) is cut to reduce the number of tokens to improve the timeliness of the vision transformer.
Figure \ref{fig:overview} depicts the architecture of our framework.
Specifically, MonoATT first adopts the DLA-34 \cite{yu2018deep} as its backbone, which takes a monocular image of size $(W \times H \times 3)$ as input and outputs a feature map of size $(W_s \times H_s \times C)$ after down-sampling with an $s$-factor.
Then, the feature map is fed into four components as follows:

\textbf{Cluster Center Estimation (CCE).}
CCE leverages a scoring network to pick out the most crucial coordinate point locations from monocular images that are worthy of being used as cluster centers based on the ranking of scores and quantitative requirements in each stage.

\textbf{Adaptive Token Transformer (ATT).}
Starting from the initial fine grid-based tokens obtained by slicing the feature map and the selected cluster centers, ATT groups tokens into clusters and merges all tokens within each cluster into one single token in each stage.
After that, a transformer network is utilized to establish a long-range attention relationship between adaptive tokens to enhance image features for Mono3D. The ATT process is composed of $N$ stages.


\textbf{Multi-stage Feature Reconstruction (MFR).}
MFR restores and aggregates all $N$ stages of irregularly shaped and differently sized tokens into an enhanced feature map of size $(W_s \times H_s \times C^{'})$.

\textbf{Monocular 3D Detection.}
MonoATT employs GUPNet \cite{lu2021geometry}, a SOTA monocular 3D object detector as its underlying detection core.

\vspace{-0.2cm}
\subsection{Cluster Center Estimation}
\label{Cluster Center Estimation}
\vspace{-0.2cm}

In order to generate adaptive tokens, it is key to be aware of the significance of each image region with respect to the Mono3D task. We have the following two observations:

\textbf{Observation 1:} \textit{As a depth knowledge, distant objects are more difficult to detect and should be paid more attention to.}

\textbf{Observation 2:} \textit{As a semantic knowledge, features of targets (\textit{e.g.}, vehicles, pedestrians, and cyclists) are more valuable than those of backgrounds, and outline features (\textit{e.g.}, lanes, boundaries, corner points) are more crucial than inner features of a target.}

Therefore, we propose to design two scoring functions to measure the depth and semantic information, respectively.
For the depth scoring function, it is straightforward to estimate the depth information using a monocular depth estimation network but it would greatly increase the computational overhead and training burden. In addition, pixel-level depth labels are required, which is not allowed in a standard Mono3D task (\emph{e.g.}, pixel-level depth labels are not available in the KITTI 3D detection dataset \cite{Geiger2012CVPR}).

Instead, we take an effective depth estimation scheme based on the camera's pinhole imaging principle and have the following proposition:

\textbf{Proposition 1:} \textit{Given the camera coordinate system $\mathbf{P}$, the virtual horizontal plane can be projected on the image plane of the camera according to the ideal pinhole camera model and the depth corresponding to each pixel on the image is determined by the camera intrinsic parameter $\mathbf{K}$.}

Particularly, we envision a virtual scene to quickly estimate the depth of each pixel in the scene, where there is a vast and infinite horizontal plane in the camera coordinate system $\mathbf{P}$.
Specifically, for each pixel locating at $(u, v)$ with an assumed depth $\hat{z}$, it can be back-projected to a point $(x_{3d}, y_{3d}, \hat{z})$ in the 3D scene:
\begin{equation}
    x_{3 d}=\frac{u-c_x}{f_x} \hat{z} \quad y_{3 d}=\frac{v-c_y}{f_y} \hat{z},
\end{equation}
where $f_x$ and $f_y$ are the focal lengths expressed in pixels along the $x-$ and $y-$ axes of the image plane and $c_x$ and $c_y$ are the possible displacements between the image center and the foot point.
These are referred to as the camera intrinsic parameters $\mathbf{K}$. 

Assume that the elevation of the camera from the ground, denoted as $H$, is known (for instance, the mean height of all vehicles in the KITTI dataset, including ego vehicles, is 1.65m \cite{Geiger2012CVPR}), the depth of a point on the depth feature map $(u, v)$ can be calculated as:
\begin{equation}
    z = \frac{f_y\cdot H}{v-c_y}.
\label{eq:z}
\end{equation}
Note that (\ref{eq:z}) is not continuous when the point is near the vanishing point, \textit{i.e.}, $v = c_y$, and does not physically hold when $v \leq c_y$.
To address this issue, we use the reciprocal to score the depth as follows:
\begin{equation}
    \mathbf{S}_{d} = -\text{ReLU}(B \frac{\mathbf{v}-c_y}{f_y\cdot H}),
\label{eq:sd}
\end{equation}
where $\mathbf{v}$ is the vector for $y-$ axis, $B$ is a constant, the ReLU activation is applied to suppress virtual depth values smaller than zero, which is not physically feasible for monocular cameras.

For the semantic scoring function, we introduce the subsequent neural network to detect the possible key points from images.
Specifically, in addition to the regular regression tasks in CenterNet \cite{zhou_objects_2019} based network, we introduce a regression branch for semantic scoring:
\begin{equation}
    \mathbf{S}_{s} = \mathbf{f}(\mathbf{H}),
\label{eq:ss}
\end{equation}
where $\mathbf{H}$ is the input image feature and $\mathbf{f}$ is the CNN architecture.
We represent the loss of point detection task as:
\begin{equation}
    \mathcal{L}_{\text{CCE}} = \text{FL}(\mathbf{g}^m(\mathbf{u_t},\mathbf{v_t}), \mathbf{S}),
\label{eq:cce}
\end{equation}
where FL is the Focal Loss used to deal with sample imbalance for key point labels; $(\mathbf{u_t},\mathbf{v_t})$ is the ground truth key point coordinate; $\mathbf{g}^m$ is the mapping function $\mathbf{g}^m:\left(\mathbb{R}^m, \mathbb{R}^m\right) \mapsto \mathbb{R}^{W_s \times H_s}$ which turns $m$ point coordinates into heatmap;
$\mathbf{S}= \mathbf{S}_{d}+\alpha \mathbf{S}_{s}$ is the score matrix for the image feature map with a size of $(W_s\times H_s)$; $\alpha$ is a hyperparameter.
The matrix addition method expands the dimensions and adds content when necessary.
The detection network is supervised by $\mathcal{L}_{\text{CCE}}$ and can be trained jointly with other Mono3D branches. 

After scoring the whole feature map, CCE calculates the mean value of the pixel scores within each token to assess the importance of that token.
We define the \textit{cluster center token} as a token that has the highest average score and serves as the starting center for token clustering.
As the number of cluster centers required for different stages is inconsistent, for stage $l$, we rank and pick out the set of cluster center tokens with number $n_l$ from $n_{l-1}$ original tokens:
\begin{equation}
    \mathbf{X}_c^l = \mathbf{g}^r(\mathbf{X}^l, n_l, \mathbf{S}),
\end{equation}
where $\mathbf{X}_c^l\in \mathbb{R}^{n_l\times C^{'}}$ is the cluster center token features; 
$\mathbf{g}^r$ is the ranking and picking function selects the highest ranked $n_l$ token features from the input tokens $\mathbf{X}^l\in \mathbb{R}^{n_{l-1}\times C^{'}}$ which is consistent with the output of previous stage $l-1$.

\begin{figure}
    \centering
    \includegraphics[width=0.85\linewidth]{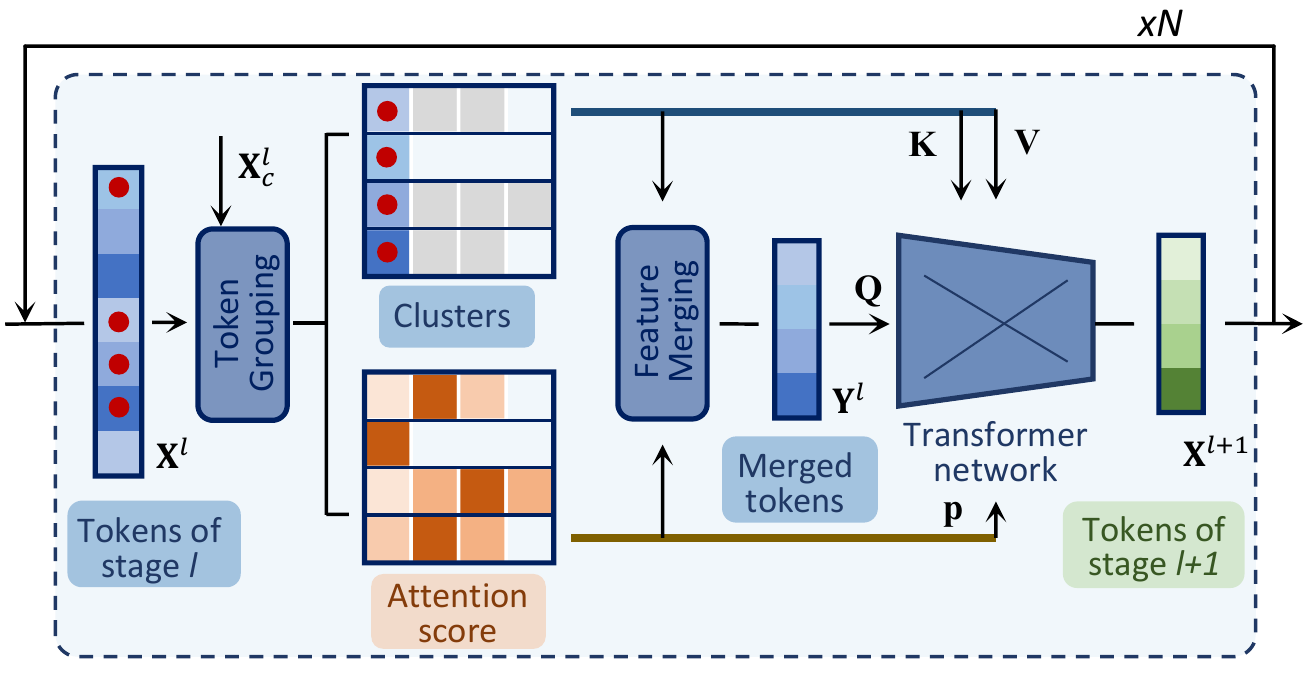}
    \caption{A schematic diagram of the ATT module.
    The pixels in the feature map are regarded as initial vision tokens.
    In each stage $l$, ATT assigns $n_{l-1}$ input tokens to the selected $n_l$ cluster center tokens $\delta$ (\textit{i.e.}, denoted as those red dots) and calculates the attention score $\mathbf{p}$ in the respective cluster.
    Then, tokens in one cluster are merged by feature $\mathbf{x}$ with score $\mathbf{p}$ to obtain a unified token $\mathbf{y}$.
    Finally, adaptive tokens are associated using a transformer and are used as the input tokens in the next stage $l+1$.
    }
\label{fig:ctm}
\vspace{-0.5cm}
\end{figure}

\vspace{-0.2cm}
\subsection{Adaptive Token Transformer}
\label{Adaptive Token Transformer}
\vspace{-0.2cm}


To enhance the image features for Mono3D, inspired by \cite{zeng2022not}, we leverage an ATT to exploit the long-range self-attention mechanism in an efficient way. As shown in Figure \ref{fig:ctm}, our AAT loops through $N$ stages, where each stage goes through two consecutive processes: \textit{i.e.}, \textit{outline-preferred token grouping}, and \textit{attention-based feature merging}.

\subsubsection{Outline-preferred Token Grouping}


It is infeasible to cluster tokens based on the straightforward spatial distance of features as it fails to identify outlines of objects due to the local feature correlation brought by 2D convolution \cite{ramamonjisoa2020predicting}.
We utilize a variant of the nearest-neighbor clustering algorithm which considers both the feature similarity and image distance between tokens \cite{zeng2022not}.

Specifically, given a set of tokens $\mathbf{X}$ and cluster center tokens $\mathbf{X}_c$, for each token, we compute the indicator $\delta_i$ as the minimal feature distance minus average pixel distance between it and any other cluster center token:
\begin{equation}
    \delta_i=\text{min}_{j:\mathbf{x}_j\in \mathbf{X}_c} (||\mathbf{x}_i-\mathbf{x}_j||_2^2 -\beta \ ||\mathbf{g}^l(\mathbf{x}_i)-\mathbf{g}^l(\mathbf{x}_j)||_2^2),
\label{eq:delta}
\end{equation}
where $\delta_i$ is the indicator that represents which cluster token $i$ should be subordinated to, $\mathbf{x}_i$ and $\mathbf{x}_j$ are feature vectors of token $i$ and $j$.
$\mathbf{g}^l$ is the look-up function that can find the mean position on the feature map corresponding to each token.
$\beta$ is a hyperparameter.
The distance constraint requires that two close tokens in the image space have to have extremely similar features in order to be clustered into the same cluster. In this way, we assign all tokens to their corresponding clusters.

\subsubsection{Attention-based Feature Merging}

To merge token features, an intuitive scheme is to directly average the token features in each cluster.
However, such a scheme would be greatly affected by outlier tokens.
Inspired by the attention mechanism \cite{rao2021dynamicvit}, we attach each token with the attention score $\mathbf{p}$ to explicitly represent the importance, which is estimated from the token features.
The token features are averaged with the guidance of attention scores as
\begin{equation}
\mathbf{y}_i=\frac{\sum_{j \in C_i} e^{p_j} \mathbf{x}_j}{\sum_{j \in C_i} e^{p_j}},
\end{equation}
where $\mathbf{y_i}$ is the merged token feature; $C_i$ indicates the set of $i$-th cluster; $\mathbf{x_j}$ and $p_j$ are the original token features and the corresponding attention score.
The region of the merged token is the union of the original cluster.

For associating adaptive tokens via the long-range self-attention mechanism, as shown in Figure \ref{fig:ctm}, merged tokens are fed into a transformer network as queries $\mathbf{Q}$, and the original tokens are used as keys $\mathbf{K}$ and values $\mathbf{V}$.
In order to differentially allow more important tokens to contribute more to the output and Reduce the impact of outliers, the attention score $\mathbf{p}$ is involved in the calculation of the attention matrix of the transformer:
\begin{equation}
    \text{Attention}(\mathbf{Q},\mathbf{K},\mathbf{V})=\text{softmax}(\frac{\mathbf{Q}\mathbf{K}^T}{\sqrt{d_k}}+\mathbf{p})\mathbf{V},
\end{equation}
where $d_k$ is the channel number of the queries.
When the dimensions of the matrix addition are inconsistent, the matrix performs expansion of the data to the appropriate dimension.
Introducing the token attention score $\mathbf{p}$ equips our ATT with the capability to focus on the critical image features when merging vision tokens.


\begin{figure}
    \centering
    \includegraphics[width=0.95\linewidth]{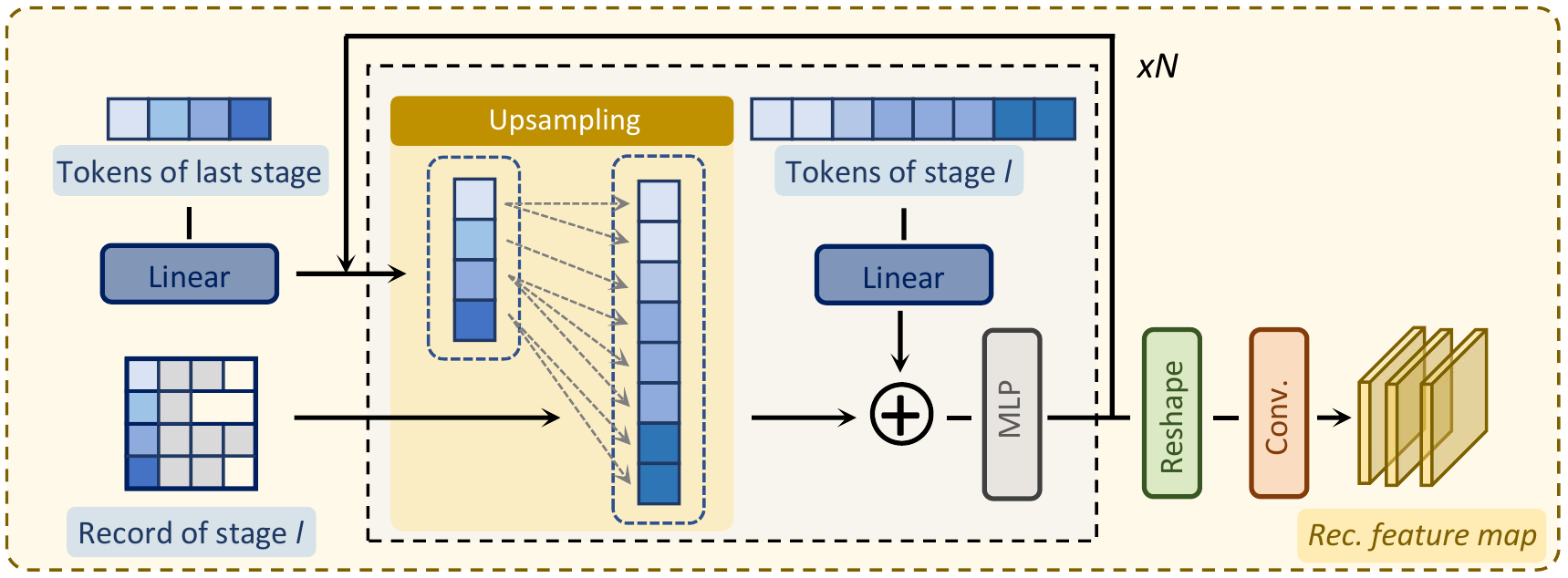}
    \caption{A illustration diagram of the MFR module.
    MFR starts from the last stage $N$ and progressively aggregates features by stacked upsampling processes and MLP blocks. In the token upsampling process, we use the recorded token relationship to copy the merged token features to the corresponding upsampled tokens. The final tokens are in one-to-one correspondence with the pixels in feature maps and reshaped to the feature maps for Mono3D.}
\label{fig:upsample}
\vspace{-0.5cm}
\end{figure}


\vspace{-0.2cm}
\subsection{Multi-stage Feature Reconstruction}
\vspace{-0.2cm}

Prior work \cite{Sun_2019_CVPR, YuanFHLZCW21} has proved the benefits of multi-stage stacking and aggregation of feature maps of different scales for detection tasks.
In order to reconstruct the feature map from irregular tokens for feature enhancement, we propose the Multi-stage Feature Reconstruction (MFR), which is able to upsample the tokens by history record and restore the feature maps.

Figure \ref{fig:upsample} shows the proposed token upsampling process.
During the token clustering and feature merging process in Section \ref{Adaptive Token Transformer}, each token is assigned to a cluster and then each cluster is represented by a single merged token.
We record the positional correspondence between the original tokens and the merged tokens.
In the upsampling process, we use the record to copy the merged token features to the corresponding upsampled tokens.
To aggregate detailed features in multiple stages, MFR adds the token features in the previous stage to the upsampled vision tokens.
The tokens are then processed by a multi-layer processing (MLP) block.
Such processing is executed progressively in $N$ stages until all tokens are aggregated.
The lowest level of tokens can be reshaped to feature maps and are processed by 2D convolution for further Mono3D detection.

Some DETR-based Mono3D detectors, such as MonoDETR \cite{zhang2022monodetr} and MonoDTR \cite{huang2022monodtr}, which use the Hungarian algorithm to detect the 3D properties of the objects directly from the tokens.

\begin{table*}[h]
\centering
\footnotesize
\begin{tabular}{l|c|c|ccc|ccc}
\toprule
\multirow{2}{*}{Method} & \multirow{2}{*}{Extra data} & \multirow{2}{*}{Time (ms)} & \multicolumn{3}{c|}{Test,\ $AP_{3D}$} & \multicolumn{3}{c}{Test,\ $AP_{BEV}$} \\ 
& & & Easy & Mod. & Hard & Easy & Mod. & Hard  \\
\midrule \midrule
PatchNet~\cite{ma2020rethinking} & \multirow{2}{*}{Depth} & 400  & 15.68 & 11.12 & 10.17 & 22.97 & 16.86 & 14.97  \\ 
D4LCN~\cite{ding2020learning} &        & 200     & 16.65 & 11.72 & 9.51  & 22.51 & 16.02 & 12.55  \\
\midrule
Kinematic3D~\cite{brazil2020kinematic}  & Multi-frames   & 120   & 19.07 & 12.72 & 9.17  & 26.69 & 17.52 & 13.10  \\
\midrule
MonoRUn~\cite{chen2021monorun}  & \multirow{2}{*}{Lidar} & 70   & 19.65 & 12.30 & 10.58 & 27.94 & 17.34 & 15.24  \\
CaDDN~\cite{reading2021categorical} &        & 630                        & 19.17 & 13.41 & 11.46 & 27.94 & 18.91 & 17.19 \\
\midrule
AutoShape~\cite{liu2021autoshape} &  CAD     & -        & 22.47 & 14.17 & 11.36 & 30.66 & 20.08 & 15.59 \\
\midrule
SMOKE~\cite{liu2020smoke} &     \multirow{5}{*}{None}   & 30                        & 14.03 & 9.76  & 7.84  & 20.83 & 14.49 & 12.75 \\
MonoFlex~\cite{zhang2021objects} &      & 30                    & 19.94 & 13.89 & 12.07 & 28.23 & 19.75 & 16.89 \\ 
GUPNet~\cite{lu2021geometry} &         & 40                     & 20.11 & 14.20 & 11.77 & -     & -     & -      \\ 
MonoDTR~\cite{huang2022monodtr} &  &  37 & 21.99 & 15.39 & 12.73 & 28.59 & 20.38 & 17.14  \\
MonoDETR~\cite{zhang2022monodetr} &  &  43 & \color{purple}{23.65} & \color{purple}{15.92} & \color{purple}{12.99} & \color{purple}{32.08} & \color{purple}{21.44} & \color{purple}{17.85}  \\
\midrule
\textbf{MonoATT~(Ours)} & None & 56 & \textbf{24.72} & \textbf{17.37} & \textbf{15.00} & \textbf{36.87} & \textbf{24.42} & \textbf{21.88}  \vspace{0.1cm}\\
\textit{Improvement} & \textit{v.s. second-best} & - &\color{blue}{+1.07} &\color{blue}{+1.45} &\color{blue}{+2.01} &\color{blue}{+4.79} &\color{blue}{+2.98} &\color{blue}{+4.03}\\
\bottomrule
\end{tabular}
\vspace{-0.3cm}
\caption{$AP_{40}$ scores(\%) of the car category on KITTI \textit{test} set at 0.7 IoU threshold referred from the KITTI benchmark website. We utilize bold to highlight the best results, and color the second-best ones and our performance gain over them in blue. Our model is ranked NO. 1 on the benchmark.}
\vspace{-0.3cm}
\label{table:perf}
\end{table*}

\begin{figure*}
    \centering
    \includegraphics[width=0.75\textwidth]{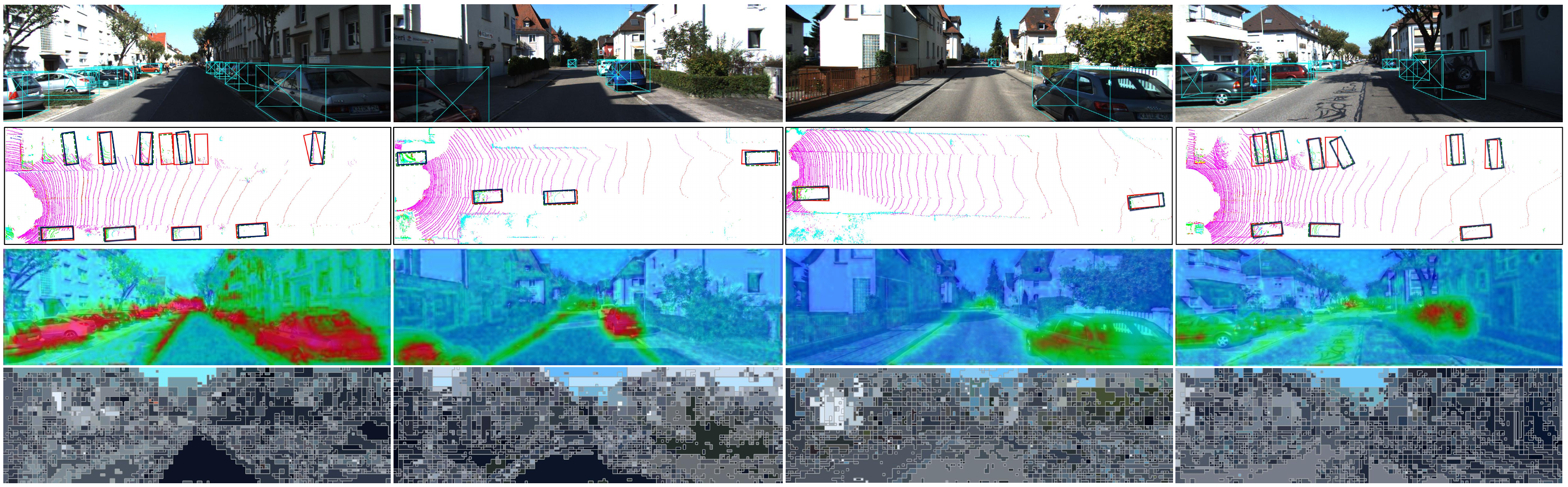}
    \vspace{-0.3cm}
    \caption{Qualitative results on KITTI dataset.
    The predicted 3D bounding boxes of our proposed MonoATT are shown in the first row. The second row shows the detection results in the bird’s eye view ($z$-direction from right to left).
    The green dashed boxes are the ground truth, and the blue and red solid boxes are the prediction results of our MonoATT and the comparison baseline (GUPNet \cite{lu2021geometry}), respectively.
    The third row and fourth rows visualize the heatmap for estimating cluster centers and adaptive tokens in the last stage.
    }
\label{fig:visual}
\vspace{-0.5cm}
\end{figure*}

\vspace{-0.2cm}
\section{Performance Evaluation}
\vspace{-0.2cm}

We conduct experiments on the widely-adopted KITTI 3D dataset \cite{Geiger2012CVPR}.
We report the detection results with three-level difficulties, \textit{i.e.} easy, moderate, and hard, in which the moderate scores are normally for ranking and the hard category is generally distant objects that are difficult to distinguish.

\vspace{-0.2cm}
\subsection{Quantitative and Qualitative Results}
\vspace{-0.2cm}

We first show the performance of our proposed MonoATT on KITTI 3D object detection benchmark \footnote{https://www.cvlibs.net/datasets/kitti/eval\_object.php?obj\_benchmark=3d} for the car category.
Comparison results with other SOTA Mono3D detectors are shown in Table \ref{table:perf}.
For the official \textit{test} set, it achieves the highest score for all kinds of samples and is ranked No.1 with no additional data inputs on all metrics.
Compared to the second-best models, MonoATT surpasses them under easy, moderate, and hard levels respectively by +1.07, +1.45, and +2.01 in $AP_{3D}$, especially achieving a significant increase (15\%) in the hard level.
The comparison fully proves the effectiveness of the proposed adaptive tokens for letting the model spend more effort on the more crucial parts of the image. 
The first two columns of Figure \ref{fig:visual} show the qualitative results on the KITTI dataset. 
Compared with the baseline model without the aid of adaptive tokens, the predictions from MonoATT are much closer to the ground truth, especially for distinct objects.
It shows that using image patches with irregular shapes and various sizes indeed helps locate the object precisely.

\vspace{-0.2cm}
\subsection{Ablation Study}
\vspace{-0.2cm}

\textbf{Effectiveness of each proposed component.}
In Table \ref{tab:component}, we conduct an ablation study to analyze the effectiveness of the proposed components: 
(a) the baseline which only uses image features for Mono3D based on GUPNet \cite{lu2021geometry};
(b) an improved version of (a) which uses a 3-stage DETR for enhancing image features with regular tokens;
(c) grouping tokens based on minimal feature distance and the token features within one cluster are averaged.
(d) the proposed \textit{outline-preferred token grouping} is exploited and token features are averaged;
(e) \textit{attention-based feature merging} is also used for token aggregation within the cluster.
All of (c), (d), and (e) do not consider the issue of how to select cluster centers, they determine them using random sampling in each stage.
Based on (e), (f) and (g) consider cluster center selection based on scores.
The difference is that (f) only uses the depth knowledge while (g) considers both the depth knowledge and the semantic knowledge.
(h) is the final version of our proposed MonoATT which additively takes into account the reconstruction of adaptive tokens into a feature map.

From a $\rightarrow$ b, it can be seen that the effectiveness of the transformer on overall performance, which helps the model to understand the long-range attention relationship between pixels.
However, the grid-based token makes the model unable to cope with small-sized targets and shows no significant change in accuracy at a distance (\textit{i.e.}, usually the hard case).
From b $\rightarrow$ c, we can observe that the use of adaptive tokens has a certain enhancement effect compared to the grid-based tokens, especially in the hard case.
Because of averaging merged token features and random sampling to select cluster centers, the performance improvement of (c) is not obvious.
From c $\rightarrow$ d, it shows that the introduced distance constraint plays a role in improving the detection accuracy.
From d $\rightarrow$ e, it can be seen that the performance gain of treating tokens in a cluster differently and using the attention mechanism to achieve feature merging.
Both (f) and (g) demonstrate that appropriate cluster centers are crucial for the generation of adaptive tokens. In addition, both the depth knowledge and the semantic knowledge are indispensable in determining cluster centers.
Group (h) indicates that for the CenterNet-based Mono3D model, transforming the token into a feature map by MFR for subsequent processing seems to be a more efficient way compared to the Hungarian algorithm in \cite{zhang2022monodetr}.

\textbf{Response time analysis.}
Other Mono3D models may require some additional operations to assist the prediction during inference. 
Compared to these methods, MonoATT is based on CenterNet \cite{zhou_objects_2019} and we adopt an efficient GPU implementation of ATT module, which only costs 9.4\% of the forward time.
We can see from Table \ref{table:perf} that our method also has a great advantage in terms of response time.

\begin{table}[t]{}
\centering
\small
\resizebox{\linewidth}{!}{
\begin{tabular}{l|c|ccc|ccc}
\toprule
 & Abla.& ATT& CCE& MFR & Easy & Mod. & Hard\\
\midrule \midrule
(a)& base. &\multirow{2}{*}{-} &\multirow{2}{*}{-} &\multirow{2}{*}{-}&22.76 & 16.46 & 13.72 \\
(b)& +T. & & & &23.18 & 17.68 & 13.95 \\
\midrule
(c)& +$\delta_i$ &\multirow{3}{*}{\checkmark} &\multirow{3}{*}{-} &\multirow{3}{*}{-}& 24.45 & 19.32 & 16.23 \\
(d)& +$\mathbf{g}^l$ & & & & 24.81 & 19.68 & 16.57\\
(e)& +$\mathbf{p}$& & & &25.62 & 20.67 & 17.76 \\
\midrule
(f)& +$\mathbf{S}_d$&\multirow{2}{*}{\checkmark} &\multirow{2}{*}{\checkmark} &\multirow{2}{*}{-} &27.72 & 21.19 & 18.15 \\
(g)& +$\mathbf{S}_s$& & & &28.36 & 21.78 & 18.87 \\
\midrule
(h)& - &\checkmark &\checkmark &\checkmark& \textbf{29.01} & \textbf{23.49} & \textbf{19.60}  
\\
\bottomrule
\end{tabular}
}
\vspace{-0.3cm}
\caption{Effectiveness of different components of our approach on the KITTI \textit{val} set for car category. The Ablation column indicates which new variables and modules we have added to the previous experimental group compared to the previous one.
\textbf{base.} is the GUPNet \cite{lu2021geometry} baseline.
\textbf{+T.} stands for the addition of a DETR-based 3-stage transformer.
}
\vspace{-0.4cm}
\label{tab:component}
\end{table}

\begin{table}[t]{}
    \centering
    \small
    \resizebox{\linewidth}{!}{
    \begin{tabular}{l|ccc|ccc}  
        \toprule
        \multirow{2}{*}{Method} & \multicolumn{3}{c|}{Val,\ $AP_{3D}$} & \multicolumn{3}{c}{Val,\ $AP_{BEV}$} \\ 
        & Easy  & Mod. & Hard & Easy & Mod.  & Hard   \\ \midrule \midrule
        MonoDTR~\cite{huang2022monodtr}  & 24.52 & 18.57 &  15.51 & 33.33 & 25.35 & 21.68 \\
         + \textbf{Ours} & \textbf{26.98} & \textbf{21.46} & \textbf{18.41} & \textbf{35.49} & \textbf{27.76} & \textbf{24.34} \\ 
        \textit{Imp.} & \color{blue}{+2.46} & \color{blue}{+2.89} & \color{blue}{+2.90} & \color{blue}{+2.16} & \color{blue}{+2.41} & \color{blue}{+2.66}  \\ \midrule
         MonoDETR~\cite{zhang2022monodetr} & 28.84 & 20.61 & 16.38 &37.86 & 26.95 &22.80 \\
          + \textbf{Ours} & \textbf{29.56} & \textbf{22.47} & \textbf{18.65} & \textbf{38.93} & \textbf{29.76} & \textbf{25.73} \\
        \textit{Imp.} & \color{blue}{+0.72} & \color{blue}{+2.13} & \color{blue}{+2.27} & \color{blue}{+1.07} & \color{blue}{+2.81} & \color{blue}{+2.93}  \\ \bottomrule        
    \end{tabular}
    }
    \vspace{-0.3cm}
    \caption{Extension of MonoATT to existing transformer-based monocular 3D object detectors. We show the $AP_{40}$ scores(\%) evaluated on KITTI3D \textit{val} set.
     \textbf{+Ours} indicates that we apply the ATT and CCE modules to the original methods. All models benefit from the MonoATT design.
    }
\label{tab:abl_equip}
\vspace{-0.6cm}
\end{table}

\textbf{Visualization of the adaptive tokens.}
To facilitate the understanding of our ATT, we visualize the cluster center scoring heatmap and the corresponding adaptive tokens in Figure \ref{fig:visual}.
In the heatmap, a warmer color means a higher score and a higher likelihood of becoming a cluster center.
As shown in the third row of the figure, the heat is concentrated on the outlines of vehicles, lane lines, and distant targets in the image.
It is worth mentioning that the outlines in the scene are of great help for depth discrimination, so even though there are no associated labels, the model can still learn that outlines are crucial for the Mono3D task based on semantics alone. 
From the visualization, we can see that the model can adaptively divide the images into tokens of different sizes and shapes.
This enables the model to use fine tokens for image parts that contain more details (\textit{e.g.}, small targets and boundary lines) and coarse tokens for image parts that do not matter (\textit{e.g.}, the sky and the ground).
This implies that ATT is able to put more experience on more important tokens, leading to more accurate predictions.

\textbf{Plugging into existing transformer-based methods.}
Our proposed approach is flexible to extend to existing transformer-based Mono3D detectors.
We respectively plug the CCE and the ATT components into MonoDTR and MonoDETR, the results of which are shown in Table \ref{tab:abl_equip}.
It can be seen that, with the aid of our CCE and ATT, these detectors can achieve further improvements on KITTI 3D \textit{val} set, demonstrating the effectiveness and flexibility of our proposed adaptive token generation. 
Particularly, MonoATT enables models to achieve more performance gains in the hard category.
For example, for MonoDETR, the $AP_{3D}$/$AP_{BEV}$ gain is +0.72/+1.07 in the easy category and +2.27/+2.93 in the hard category.

\textbf{Efficacy for detecting objects at different distances.}
In Table \ref{tab:abl_distance}, we compare the accuracy gain of our model for detecting objects at different distances.
We present the accuracy (\%) of ours with Kinemantic3D, MonoDTR, and MonoDETR as the baselines in the KITTI \textit{val} set for the following three distance ranges: near (5m-10m), middle (20-25m), and far (40m-45m). 
It can be seen that MonoDTR and MonoDETR using a transformer with grid-based tokens do not perform well in the far case, although they outperform Kinemantic3D in terms of overall accuracy.
For MonoDTR, the $AP_{3D}$/$AP_{BEV}$ gain is +8.79/+8.36 on the far case.
From this, we can see that our method has a significant improvement in detecting distant objects.

\begin{table}[]{}
    \centering
    \small
    \resizebox{\linewidth}{!}{
    \begin{tabular}{l|ccc|ccc}  
        \toprule
        \multirow{2}{*}{Method} & \multicolumn{3}{c|}{Val,\ $AP_{3D}$} & \multicolumn{3}{c}{Val,\ $AP_{BEV}$} \\ 
        & Near  & Mid. & Far & Near  & Mid. & Far   \\ \midrule \midrule
        Kinematic3D ~\cite{brazil2020kinematic}  & 34.52 & 16.50 &  5.82 & 246.86 & 22.81 & 7.35 \\
         + \textbf{Ours} & \textbf{35.70} & \textbf{21.86} & \textbf{10.48} & \textbf{48.23} & \textbf{27.84} & \textbf{12.34} \\ 
        \textit{Imp.} & \color{blue}{+1.18} & \color{blue}{+5.36} & \color{blue}{+4.66} & \color{blue}{+1.37} & \color{blue}{+5.03} & \color{blue}{+4.99}  \\ \midrule
        MonoDTR~\cite{huang2022monodtr}  & 48.51 & 17.87 &  2.16 & 61.25 & 25.03 & 3.29 \\
         + \textbf{Ours} & \textbf{49.69} & \textbf{22.49} & \textbf{10.95} & \textbf{63.24} & \textbf{29.81} & \textbf{11.65} \\ 
        \textit{Imp.} & \color{blue}{+1.18} & \color{blue}{+4.62} & \color{blue}{+8.79} & \color{blue}{+1.99} & \color{blue}{+4.78} & \color{blue}{+8.36}  \\ \midrule
         MonoDETR~\cite{zhang2022monodetr} & 48.66 & 17.91 & 2.35 &61.21 & 25.25 &3.38 \\
          + \textbf{Ours} & \textbf{49.92} & \textbf{22.48} & \textbf{11.07} & \textbf{63.29} & \textbf{30.00} & \textbf{11.91} \\
        \textit{Imp.} & \color{blue}{+1.26} & \color{blue}{+4.57} & \color{blue}{+8.72} & \color{blue}{+2.08} & \color{blue}{+4.75} & \color{blue}{+8.53}  \\ \bottomrule        
    \end{tabular}
    }
    \vspace{-0.3cm}
    \caption{
    The comparison of the performance gain of our MonoATT over existing models at different distances.
    We show the $AP_{40}$ scores(\%) evaluated on KITTI3D \textit{val} set.
     \textbf{+Ours} indicates that we apply our modules to the original methods. 
     Near (5m-10m), middle (20m-25m), and far (40m-45m) are three different distance intervals.
     All models benefit from the MonoATT design, especially for far objects.
    }
\vspace{-0.6cm}
\label{tab:abl_distance}
\end{table}

\vspace{-0.2cm}
\section{Conclusion}
\vspace{-0.2cm}

In this paper, we have proposed a Mono3D framework, called \emph{MonoATT}, which can effectively utilize the generated adaptive vision tokens to improve online Mono3D. 
The advantages of MonoATT are two-fold: 1) it can greatly improve the Mono3D accuracy, especially for far objects, which is an open issue for Mono3D; 2) it can guarantee low latency of Mono3D detectors by omitting backgrounds suitable for appealing mobile applications. 
Nevertheless, MonoATT still has two main limitations as follows: 
1) the computational complexity of the nearest neighbor algorithm in stage 1 is still linear with respect to the token number, which limits the speed of MonoATT for a large initial token input;
2) it heavily counts on the scoring network in CCE, which may sometimes be disturbed by semantic noise, such as complex vegetation areas.
These limitations also direct our future work. We have implemented Mono3D and conducted extensive experiments on the real-world KITTI dataset. MonoATT yields the best performance compared with the SOTA methods by a large margin and is ranked number one on the KITTI 3D benchmark.

\vspace{-0.2cm}
\section*{Acknowledgement}
\vspace{-0.2cm}

This research was supported in part by National Natural Science Foundation of China (Grant No. 61972081) and the Natural Science Foundation of Shanghai (Grant No. 22ZR1400200).

{\small
\bibliographystyle{ieee_fullname}
\bibliography{egbib}
}

\end{document}